\documentclass[12pt]{article}
\usepackage[utf8]{inputenc}
\usepackage{times}
\usepackage{setspace}
\usepackage[margin=1in]{geometry}
\usepackage{titlesec}
\usepackage{enumitem}

\usepackage{url}
\usepackage[numbers]{natbib}
\usepackage[colorlinks=true, linkcolor=blue, citecolor=blue]{hyperref}

\usepackage{graphicx}

\titleformat{\section}{\bfseries\large}{\thesection.}{1em}{}
\titleformat{\subsection}{\bfseries\normalsize}{\thesubsection.}{1em}{}

\title{Humanizing AI Grading: Student-Centered Insights on Fairness, Trust, Consistency and Transparency }
\author{
Bahare Riahi\thanks{Computer Science, North Carolina State University, Raleigh, NC, USA. Email: \texttt{briahi@ncsu.edu}. ORCID: 0009-0005-4560-4857} \\
\and
Viktoriia Storozhevykh\thanks{Computer Science, North Carolina State University, Raleigh, NC, USA. Email: \texttt{vstoroz@ncsu.edu}. ORCID: 0009-0002-3453-7781} \\
\and
Veronica Catet\'e\thanks{Computer Science, North Carolina State University, Raleigh, NC, USA. Email: \texttt{vmcatete@ncsu.edu}. ORCID: 0000-0002-7620-7708} 
}

\date{}
\begin{document}
\maketitle

\begin{abstract} 

This study investigates students’ perceptions of Artificial Intelligence (AI) grading systems in an undergraduate computer science course (n = 27), focusing on a block-based programming final project. Guided by the ethical principles' framework articulated by Jobin (2019), our study focused on fairness, trust, consistency and transparency in AI grading, using AI-generated feedback as a second version compared to the original human-graded feedback. Findings reveal concerns about AI’s lack of contextual understanding and personalization. We recommend that equitable and trustworthy AI systems reflect human judgment, flexibility, and empathy, serving as supplementary tools under human oversight. This work contributes to ethics-centered assessment practices by amplifying student voices and offering design principles for humanizing AI in designed learning environments.

\end{abstract}

\section{Objectives}

With the development of AI, numerous ethical concerns have evolved regarding its usage. Concerns in various cases such as accidents in self-autonomous vehicles \cite{banks2018driver}, algorithmic bias leading to discrimination \cite{zou2018ai}, breaches of user privacy \cite{lee2024deepfakes} highlighted the potential risks. To mitigate risks, ethical factors should be investigated from users' perspectives and finally addressed in AI tools.
There are various opinions about AI's role in education; however, there is limited empirical research on students' perceptions of using AI and learning \cite{tossell2024student}. While implementation may vary by model (e.g., FairLearn), fundamental principles like fairness, transparency, consistency and trust \cite{jobin2019global} are crucial in education where students’ sense of equity is shaped by AI's decisions. Since successful AI integration is highly dependent on human trust \cite{glikson2020human}, it is essential that students believe in the precision of AI-generated grades and feedback. Without this trust, their learning may be hindered \cite{jonall2024artificial, tossell2024student}. Although large language models can effectively assess open-ended questions \cite{pinto2023large}, challenges remain regarding reliability and fairness in assessment and using rubrics \cite{xie2024grade}. However, manual grading is also prone to inter-rater variability and unfairness, making AI an overall more consistent and efficient alternative for grading \cite{jamil2023toward,yousef2025begrading,hickman2024automated}.
AI increases students' autonomy and supports learning by providing personalized feedback, offering a more sustainable and consistent alternative to traditional methods like peer review \cite{rudolph2023chatgpt, alipour2024improving}. Some students also perceive AI-based assessment as more objective, as it minimizes personal bias in grading \cite{jonall2024artificial}. These factors highlight the importance of fairness as a core principle in the design and implementation of AI learning systems.
In the case of block-based programming, many instructors have found manual grading to be challenging due to difficulties with understanding the code and overwhelming number of submissions \cite{milliken2021exploring}. Therefore, teachers have indicated the need for automated grading to reduce the time required to assess all assignments \cite{riahi2025comparative}. 

This paper explores students' perceptions of AI-assisted assessment in block-based programming class projects, focusing on ethical factors taken from Jobin (2019) framework.
Our research questions are:

\begin{itemize}
  
    \item \textbf{RQ1:} Do students perceive feedback generated by the AI-grading system as transparent and understandable compared to human graders?
      \item \textbf{RQ2:} Do students perceive feedback generated by the AI-grading system as fair and accurate compared to human graders?
    \item \textbf{RQ3:} Do students perceive that AI-grading systems provide as consistent and equitable feedback on their projects compared to human grader?
    \item \textbf{RQ4:} Do students trust AI-generated feedback, and prefer it over human feedback?
    \item \textbf{RQ5:} What improvements are recommended to humanize AI grading systems and enhance their acceptance among students?
\end{itemize}

\section{Theoretical Framework}
This study uses the ethical principles framework articulated by Jobin (2019) to examine students’ perceptions of AI-assisted grading in a block-based programming context. Jobin et al. conducted a meta-analysis of 84 global AI ethics guidelines and identified five commonly recurring principles: transparency, justice/fairness, non-maleficence, responsibility, and privacy. For this study, we focus on three of these: fairness, trust, transparency plus consistency, since they are most relevant to how students interpret, trust, and respond to AI-generated feedback on their class projects.
Trusting AI entails the user believing that the model is capable of autonomously executing tasks the user asks of it, in addition to the user perceiving the technology as useful and reliable in providing accurate information \cite{glikson2020human}. People can be reluctant in trusting AI, owing to the novelty, but if their perception of AI is changed to be credible, fair, and transparent, that trust can greatly increase \cite{ bach2024systematic}. Consistency refers to the degree to which similar judgments are made across assessors—such as between AI and human graders—based on the same student work \cite{donaldson2012systematic, guskey2024addressing}.
A fair AI model is defined as one that performs consistently across diverse users and groups and is developed through ethical and transparent research practices \cite{ryan2023integrating}. 
Transparency in learning systems concerns how clearly AI-generated feedback is presented and understood \cite{glikson2020human}. Students must know that the feedback is produced by AI, how the system operates, and how accurate or robust it is \cite{hofman2023transparency}, especially to foster trust and learning \cite{smith2024code}.

\section{Methods, Modes of Inquiry and Data Sources}

This study used undergraduate final project submissions from a computer science course to evaluate projects using AI (ChatGPT team version, licensed by our institution). 
These projects had previously been graded by the course's teaching assistants (TAs), and students had already received feedback. We evaluated the projects using the same rubric that the TAs had used for grading, and we also uploaded the course description to the AI system.
This study has been approved by the Institutional Review Board (IRB). To ensure compliance with FERPA, we asked students for consent through a form to use their submissions for research. Additionally, to avoid using the submissions for AI training, we used ChatGPT Team version to generate the feedback, and the TAs anonymized the files. Once the AI-generated feedback was ready, the TAs sent it to the students along with a survey. The survey consisted of three parts:

\begin{itemize}
    \item \textbf{Students' Perception:} The first part collected students’ opinions on the AI-generated feedback using a 5-point Likert scale \cite{russo2021differences}.
    \item \textbf{Evaluation of Feedback and comparison:} The second part included 5-point Likert scale questions alike, assessing the clarity, transparency, fairness, transparency of the feedback, and comparison to human (TA) generated feedback.
    \item \textbf{Open-Ended Questions:} The final part included open-ended questions, asking students for their thoughts on the experience, preference and if they noticed any differences between the two versions of feedback. It also asked whether they liked or disliked the AI feedback. Additionally, students were asked about their grades in both assessment versions to ensure that their responses were not influenced by receiving a higher grade.
\end{itemize}
 We analyzed the data of the open-ended questions by coding the responses and using thematic analysis \cite{alhojailan2012thematic,maguire2017doing}.
To minimize bias, some questions were asked in an opposite direction, asking to what extent they agree or disagree if the TAs' feedback was clearer, ensuring a balanced comparison. 

\section{Results and Conclusions}
In our survey data and analysis, we gained insights into students' perceptions of their experience with AI-based assessment of their class projects. Our results are from both quantitative (survey) and qualitative data (open-ended questions). 

\subsection{Quantitative Data Analysis}
The first two sections of our survey were quantitative, with responses assigned numeric weights to calculate each section's score, that was ranging from strongly agree to strongly disagree. The first section measured students' perceptions of AI-generated feedback on fairness, transparency, clarity, and trust using a Likert scale. Results showed that students found the grading fair (mean score: 4.2), clear and transparent (4.12), consistent (4.08), and reported high trust in the system (4.03).

Section 2 compared AI-generated feedback with that provided by TAs. While 60\% of students believed the TA feedback was fairer, despite receiving higher grades from AI, 57\% found the AI-generated feedback to be more transparent and clearer. Additionally, 63\% of students perceived consistency between AI and TA feedback, though 55\%  reported greater trust in TA feedback.

\subsection{Qualitative Data Analysis}
In the open-ended questions, students were asked to reflect on their overall experiences, concerns, and preferences regarding the two versions of grading. Their responses highlighted perceived differences in fairness, clarity, transparency, and consistency between the AI-graded and human-graded feedback.
To analyze our qualitative data, we conducted thematic analysis to identify recurring patterns and insights regarding participants’ perceptions of AI versus human grading. Using an inductive coding approach, responses were reviewed line-by-line and grouped into overarching themes such as clarity, fairness, consistency, and trust. Each theme was further broken down into sub-themes. This qualitative analysis allowed us to capture nuanced opinions and provide deeper context to the quantitative findings.
We extracted four main themes and related sub-themes, as follow:

In theme 1, Clarity and Transparency, students expressed mixed feelings about AI and TA grading. In sub-theme 1.1, many students found AI feedback clearer and appreciated the clear structure, because they received separate notes for each category. However, Sub-theme 1.2 revealed a preference for TA feedback grounded in human judgment and situational understanding, therefore, students felt AI grading lacked transparency, particularly in explaining why certain criteria were met or missed. 

In theme 2, Fairness and Accuracy, sub-theme 2.1: AI Missed Key Elements or Was Inaccurate, revealed leaks in the accuracy, and some students believed the AI system missed important components of their work, and they expressed concerns about accuracy. In other words, they felt TA grading is fairer and more nuanced, with accountable evaluations, whereas AI grading was perceived less precise and sometimes missed key aspects. Sub-theme 2.2 shows AI is fairer since deducted points is more associated with mistakes.

Addressing Theme 3, perceived grading consistency, Sub-theme 3.1 focused on inconsistencies in AI grading. Some students felt that AI grading lacked consistency with TA grading, often missing key elements, grading based on surface-level features, and lacking the contextual understanding provided by human graders, such as TAs. However, Sub-theme 3.2 highlighted the consistency in both AI and TA grading, with students perceiving AI as consistent due to receiving similar evaluations to those from TAs.

Regarding Theme 4, trust and preference, students expressed varied views. Some students preferred human insights due to AI's potential errors and inconsistencies. Some trusted AI for its clarity and transparency and others, expressed no strong preference between AI and TA grading.

Finally, our results show that, in response to \textbf{RQ1}, students found the AI-generated feedback transparent and easy to understand, often highlighting the clarity and structured explanations provided and reasoning behind it. They expressed more clarity in AI feedback and identical transparency in both versions. To answer \textbf{RQ2}, students generally viewed the AI feedback as moderately fair but believe TA feedback was fairer.

Answering \textbf{RQ3}, our data demonstrated nearly unanimous agreement on existing more consistency in the AI-grading system, compared to teachers' feedback.
Addressing \textbf{RQ4}, participants indicated a moderate level of trust due to clarity, consistency and reliability to AI they trust and prefer human grading feedback more than AI.

\subsection{Recommendation for Humanizing AI Systems in Education In future}
This study contributes to the growing understanding of AI-assisted assessment by highlighting that while AI promotes efficiency and scalability, it also presents important challenges that must be addressed in the future.
Answering \textbf{RQ5}, our findings suggest that to make AI grading systems more equitable and acceptable to students, they must be humanized. This includes integrating AI-generated feedback with human oversight, incorporating interpretive judgment, and reflecting the contextual understanding and personalization that human graders provide. AI should be a supportive tool, not a replacement for human graders, allowing educators to modify and combine AI feedback with their own insights. Additionally, AI feedback should adopt an empathetic, constructive, and encouraging tone, mirroring how teachers communicate. When used collaboratively, AI can help scale education while ensuring fairness and support for all students, especially those from marginalized backgrounds.

\section{Scientific significance of the study or work}
AI grading tends to be more focused on technical guidelines such as syntax and giving explanatory feedback for general improvement. Many students still prefer human-grading because it focused narrowly on the overall code quality and overlooked code readability, students' effort and course progression. In one of the examples, the AI and the TA had different interpretations of what constituted a "major mechanic" and evaluated code quality with different standards and in another example, the AI gave full points in some areas where the human grader gave fewer points, and vice versa. These differences suggest that AI may benefit from better alignment with human evaluative standards.

\section{Future Work and Limitations}
The size of our sample size may restrict the generalizability of our findings. We plan to extend this research by conducting similar studies in additional undergraduate courses during Fall 2025 at our institution. This will allow us to collect more extensive qualitative and quantitative data and to examine student perceptions across different course contexts and projects. 
\newpage

 \begin{figure}[h!]
    \centering
    \includegraphics[width=0.7\columnwidth]{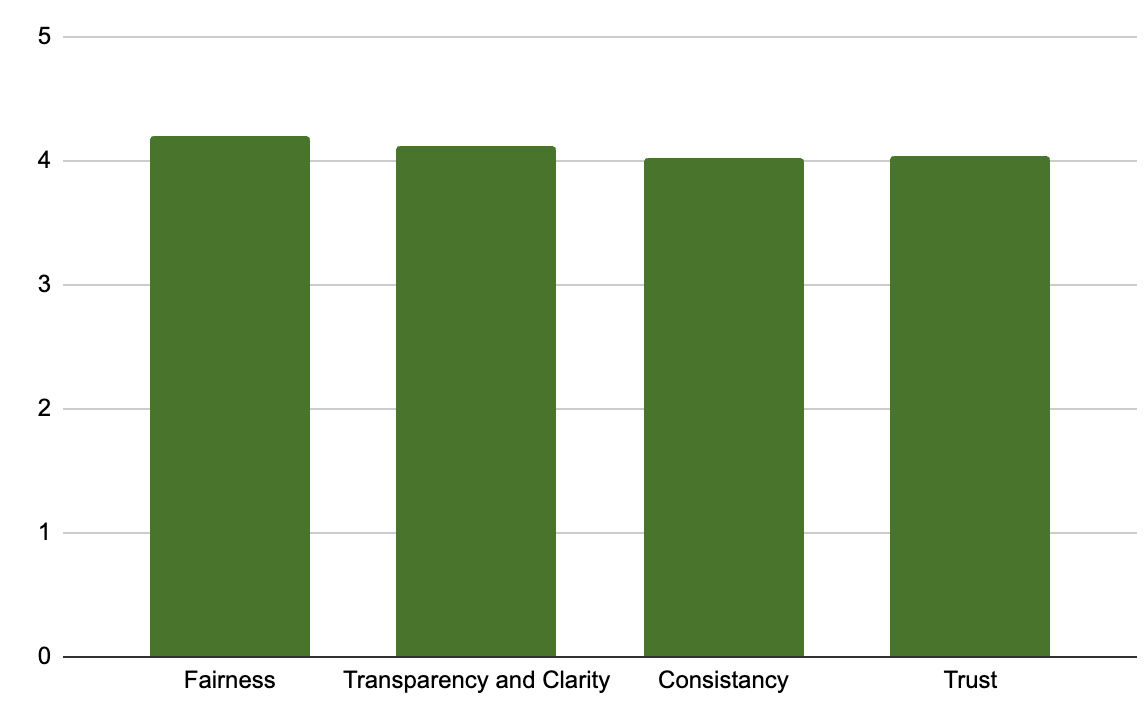}
    \caption{ Section 1 of survey: Mean Scores for Perceived Fairness, Clarity, Transparency, and Consistency in AI-Graded Feedback}
\end{figure}

 \begin{figure}[h!]
    \centering
    \includegraphics[width=0.7\columnwidth]{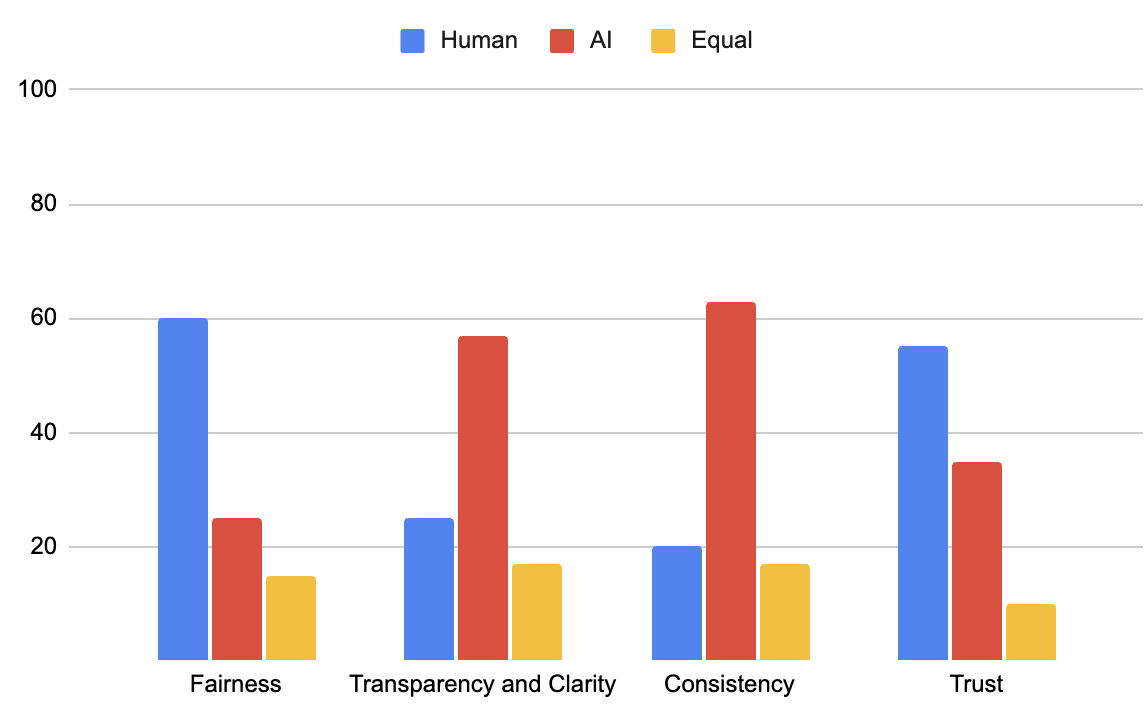}
    \caption{Section 2 of survey: Comparison of Perceived Grading Qualities Across AI, Human, and Equal Versions}
\end{figure}

 \begin{figure}[h!]
    \centering
    \includegraphics[width=1.1\columnwidth]{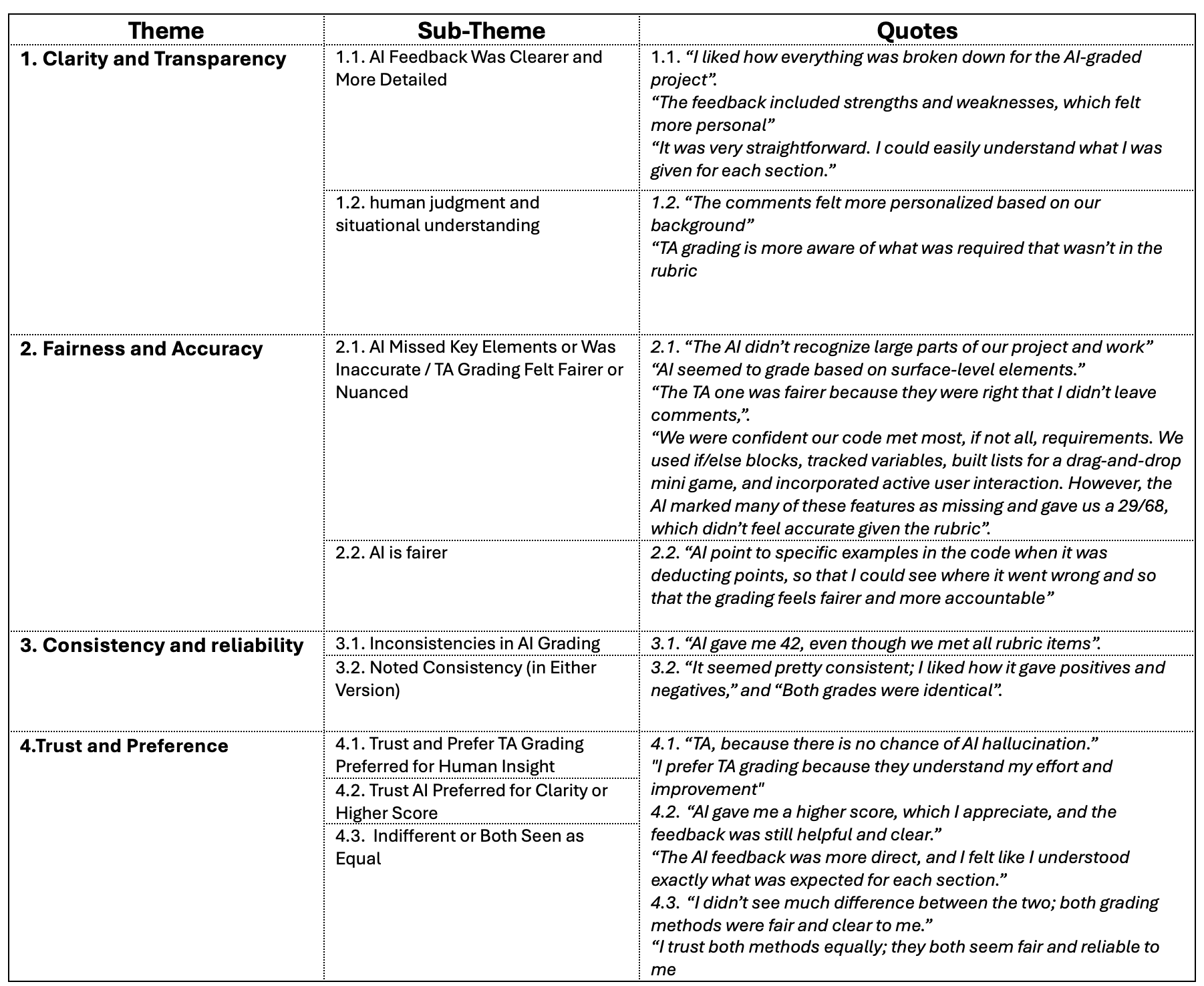}
    \caption{Themes, Sub-themes and Quotes in qualitative analysis (Thematic analysis)}
\end{figure}

\newpage
\bibliographystyle{ACM-Reference-Format}
\bibliography{aera_refs.bib}

\end{document}